%% file: acl_latex_v3.tex
\pdfoutput=1

\documentclass[11pt]{article}

\usepackage[final]{acl}
\usepackage{tabularx}
\usepackage{booktabs}
\usepackage{multirow}
\usepackage{times}
\usepackage{latexsym}
\usepackage{amsmath}
\usepackage{algorithm}
\usepackage{algorithmic}
\usepackage[T1]{fontenc}

\usepackage[utf8]{inputenc}

\usepackage{microtype}

\usepackage{inconsolata}

\usepackage{xurl}

\usepackage{graphicx}

\usepackage[T2A,T1]{fontenc}   
\usepackage[bulgarian,english]{babel} 

\title{Kinship Data Benchmark for Multi-hop Reasoning}

\author{Tianda Sun \\
  Department of Computer Science \\
  University of York, UK \\
  \texttt{tianda.sun@york.ac.uk} \\\And
  Dimitar Kazakov \\
  Department of Computer Science \\
  University of York, UK \\
  \texttt{dimitar.kazakov@york.ac.uk} \\}

\usepackage{enumitem}

\setlist[itemize]{itemsep=2pt,topsep=2pt,parsep=0pt}

\begin{document}
\maketitle
\begin{abstract}
Multi-hop kinship reasoning is a natural testbed for LLM compositionality, but existing benchmarks (notably CLUTRR) cover only the descriptive Eskimo system. We introduce KinshipQA, a procedurally-generated benchmark covering seven anthropologically-documented kinship systems (Eskimo, Sudanese, Hawaiian, Iroquois, Dravidian, Crow, Omaha) and up to six reasoning hops, with a tunable simulator horizon that eliminates exact-instance pretraining overlap. Evaluating six LLMs, we find a 40.9\% accuracy drop when reasoning shifts from biological multi-hop to culturally-marked classification on the five non-descriptive systems. The drop holds for every non-descriptive system and is largest for the two skewing systems (Crow, Omaha), persists under chain-of-thought and few-shot prompting, and compounds with depth: at 5--6 hops cultural override falls to 10.6\% while biological composition over the same chains remains at 58.6\%. Under identical rule access humans reach 89.0\% versus 50.7\% for LLMs, so the questions are reliably solvable once the rule is supplied. Two follow-up experiments suggest distinct contributors. A fictional-rule control swapping system labels and kin terms for invented strings raises accuracy by 6.1\%, implicating familiar English surface forms. An in-context-rule probe prepending the override rule helps skewing systems ($+17.1$\%) but \emph{hurts} non-skewing systems whose baseline already exceeds $\sim$60\% ($-13.4$\%), consistent with a missing skewing prior alongside rule interference where the model already has a working approximation. Our code and data are publicly available on \href{https://github.com/TiandaSun/Kinship-Data-Benchmark-for-Multi-hop-Reasoning}{GitHub}.
\end{abstract}

\section{Introduction}

\begin{figure}[t]
    \centering
    \includegraphics[width=1.12\columnwidth]{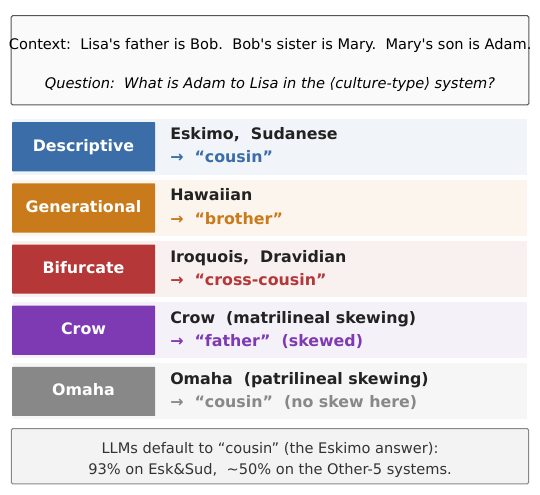}
    \caption{KinshipQA cultural disambiguation: the same kinship configuration (Lisa's father's sister's son Adam) maps to four distinct cultural answers across the typological classes; Crow's matrilineal skewing reclassifies Adam as ``father'' while Omaha's patrilineal skewing leaves this position unchanged. LLMs default to the descriptive Eskimo answer (``cousin''): 93\% on Eskimo+Sudanese (hereafter Esk\&Sud), $\sim$50\% on the Other-5 systems.}
    \label{fig:headline}
\end{figure}

\input{latex/introduction-Dimitar}

\section{Related Work}

\paragraph{Multi-hop reasoning benchmarks.} Multi-hop reasoning remains a fundamental challenge for LLMs \cite{qiao_reasoning_2023}. HotpotQA~\cite{yang_hotpotqa_2018} and 2WikiMultihopQA~\cite{ho_constructing_2020} face training-data contamination and lack reasoning-chain verification; MuSiQue~\citep{trivedi_musique_2022} mitigates the latter via single-hop composition under a connectivity constraint, while MuSR~\citep{sprague_musr_2024} pushes multi-step reasoning into narrative domains. Recent benchmarks add structural demands or contamination resistance (MoreHopQA, \citealp{schnitzler_morehopqa_2024}; CofCA, \citealp{wu_cofca_2025}; CompoST, \citealp{schmidt_compost_2024}), and domain-specific extensions (BioHopR, \citealp{kim_biohopr_2025}; FamilyTool, \citealp{luo_familytool_2025}) show that hop count alone is incomplete. Rule-application benchmarks supply rules explicitly (RuleArena, \citealp{zhou_rulearena_2025}; \citealt{thota_semantic_override_2026}); our main task instead names but does not state the cultural rule, with the in-context-rule probe (App.~\ref{app:rule_context}) comparing both regimes. \citet{li_understanding_2024} localise multi-step failures to middle-layer self-attention.
\paragraph{Kinship reasoning benchmarks.} There is a history of research using kin data as a machine learning testbed. Earlier knowledge-graph reasoning work used the standard Kinship multi-relational benchmark for path-based reinforcement learning~\citep{lin_multihop_2018}, predating the LLM-era kinship benchmarks discussed below. CLUTRR \cite{sinha_clutrr_2019} uses kinship to evaluate compositional reasoning: from semi-synthetic stories (\emph{Alice is Bob's mother; Bob is Carol's father}), models must infer unstated relations (\emph{Alice is Carol's grandmother}). CLUTRR tests \textit{rule induction} over a single (descriptive Eskimo) kinship typology. Two recent works extend kinship reasoning beyond CLUTRR's single-typology setting but along orthogonal axes to ours: \citet{zhang_path_2024} introduce a Chinese kinship reasoning dataset alongside CLUTRR as one of four evaluation domains, spanning a single descriptive-type system with richer maternal/paternal lexicalisation than English; and \citet{lian_lingbench_2025} compile LingBench++, a cross-cultural kinship-term puzzle set drawn from the International Linguistics Olympiad archive, covering more than 90 low-resource languages as a static rule-induction benchmark in the IOL paradigm. Our work differs along three axes: we cover Morgan's seven major classification systems selected for typological coverage rather than language coverage; our questions are procedurally generated from RDF/OWL genealogies under formal culture-specific marriage-rule constraints, with a tunable simulator horizon that eliminates exact-instance overlap with pretraining; and we test rule \textit{application} given a named cultural system and an in-context override cue, separating it from rule induction in part via a fictional-rule control (\S\ref{sec:interventions}).

\paragraph{Cultural knowledge in NLP.} Cross-cultural NLP has gained attention as researchers recognise that language technologies must account for cultural variation, with kinship identified as an under-researched category where evaluation data are lacking \cite{liu_culturally_2024}; existing cultural benchmarks have been criticised for reducing culture to static facts \cite{alkhamissi_hire_2025}. Unlike biological kinship (universal facts of reproduction), cultural kinship operates as a well-defined ``constructed, computational system'' with formal rules and anthropologically-documented variation \cite{read_cultural_2012}, making it a natural testbed for multi-relational reasoning.


\section{Methodology}
\subsection{Task Formulation}

We frame kinship reasoning as a reading comprehension task. Given a natural language context $\mathcal{C}$ describing family relationships and a question  $\mathcal{Q}$ about a specific individual, the model must 
produce an answer  $\mathcal{A}$ that is either a single entity or a set of entities. Formally:   $f(\mathcal{C}, \mathcal{Q}) \to \mathcal{A}$ .

The key challenge lies in distinguishing between two reasoning modes: biological reasoning, which follows 
universal genealogical facts (e.g., ``mother's sister'' → ``aunt''), and cultural-rule application, which 
requires deploying culture-specific classification rules that may override biological defaults 
(e.g., in Hawaiian kinship, ``mother's sister'' → ``mother''). Our benchmark tests both capabilities 
across controlled complexity levels measured by the number of relationship hops (n-hops) required 
to derive the answer.

\subsection{Kinship Systems as Test Domains}

We select kinship systems as our test domain for three reasons. First, kinship provides naturally 
controlled complexity: relationships form tree structures where reasoning depth (n-hops) is 
precisely measurable. Second, kinship systems exhibit well-documented cultural variation with 
formal, rule-based classifications—ideal for systematic evaluation. Third, the domain enables 
procedural generation of novel instances, eliminating exact-instance overlap with pretraining data.

\begin{table}[t]
\centering
\caption{Overview of seven kinship classification systems in KinshipQA. F=Father, FB=Father's Brother, FZS=Father's Sister's Son, MB=Mother's Brother, MBS=MB's Son. See Appendix~\ref{app:kinship-systems} for full anthropological descriptions and per-system override rules.}
\small
\begin{tabular}{lccl}
\toprule
\textbf{System} & \textbf{Type} & \textbf{Key Rule} \\
\midrule
Eskimo & Descriptive  & F $\neq$ FB \\
Sudanese & Descriptive  & All terms unique \\
Hawaiian & Generational  & F = FB \\
Iroquois & Bifurcate  & Parallel $\neq$ Cross \\
Dravidian & Bifurcate  & Cross = Spouse \\
Crow & Mat. Skewing  & FZS = F \\
Omaha & Pat. Skewing  & MBS = MB \\
\bottomrule
\end{tabular}
\label{tab:systems}
\end{table}


Our seven systems span Morgan's descriptive--classificatory dichotomy~\cite{morgan_systems_1871} as refined by Murdock~\cite{murdock_social_1949} into the six-fold classification (Eskimo, Hawaiian, Iroquois, Crow, Omaha, Sudanese), plus the Dravidian elaboration of Trautmann~\cite{trautmann_dravidian_1981}. Table~\ref{tab:systems} summarises them; full anthropological descriptions are in App.~\ref{app:kinship-systems}.

\subsection{Benchmark Generator Pipeline}

\begin{figure*}
    \centering
    \includegraphics[width=1\linewidth]{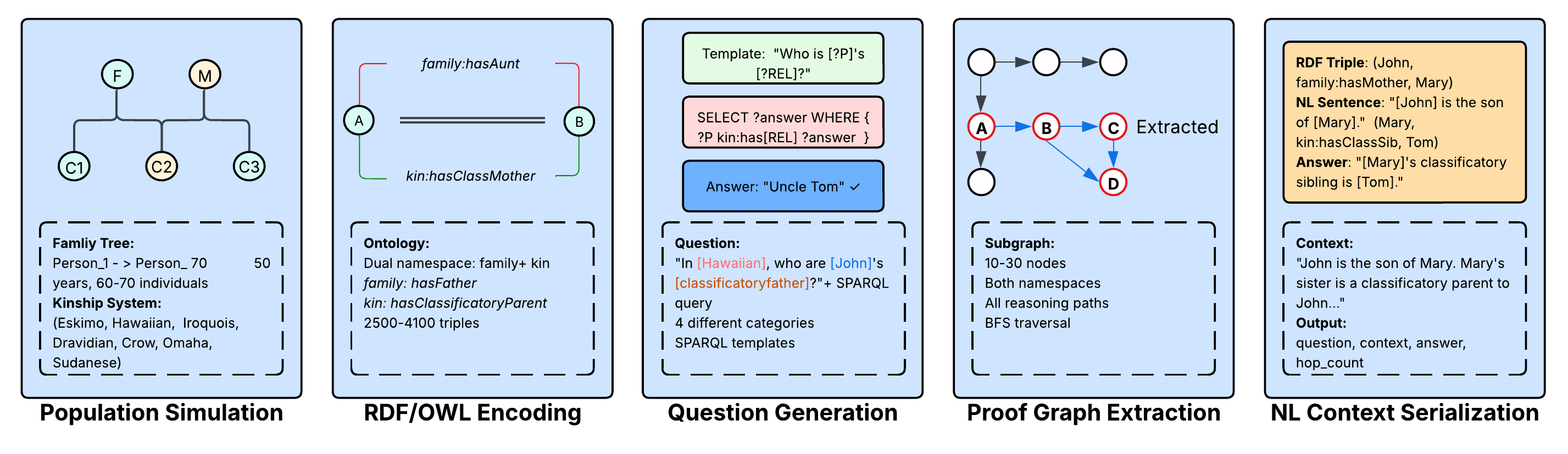}
    \caption{KinshipQA generation pipeline. (1) \textbf{Population Simulation} --- multi-generational family trees under culture-specific marriage rules. (2) \textbf{RDF/OWL Encoding} --- dual namespaces (\texttt{family:} biological; \texttt{kin:} cultural). (3) \textbf{Question Generation} --- four categories at controlled hop depths. (4) \textbf{Proof Graph Extraction} --- minimal reasoning subgraph per question. (5) \textbf{NL Context Serialisation} --- RDF subgraph $\to$ NL for the LLM.}
    \label{fig:pipeline}
\end{figure*}

Figure~\ref{fig:pipeline} illustrates our five-stage, fully automated pipeline. We simulate multi-generational family trees under culture-specific marriage constraints over tunable horizons (50-year baseline yields 60--70 individuals; 150/200-year extensions yield 60--360 individuals depending on system marriage rules)---e.g., Eskimo prohibits only sibling marriage; Iroquois allows cross- but not parallel-cousin marriage; Dravidian prefers cross-cousin marriage; Crow and Omaha enforce clan-based prohibitions under matrilineal and patrilineal descent. Trees are encoded as RDF with a dual-namespace ontology---\texttt{family:} for biological relationships and \texttt{kin:} for cultural classifications---enabling precise evaluation of whether models distinguish biological facts from cultural categories. Questions are generated by mapping biological relationship paths (e.g., mother $\rightarrow$ sister $\rightarrow$ child) to system-specific kinship terms, ensuring genuine multi-hop reasoning rather than memorised labels; ground truth is computed by SPARQL queries against the ontology, and the minimal sub-graph supporting each question is serialised to natural language by templates (e.g., ``Alice is the mother of Bob''). The benchmark therefore has a machine-checkable closure: a deterministic SPARQL executor over the released RDF/OWL ontology returns the gold answer on 100\% of the 3{,}931 questions by construction, so the LLM gaps reported below arise from natural-language interpretation and reasoning over the supplied facts rather than gold inconsistency or evaluator ambiguity (the natural-language surface realisation is validated separately by the human baseline in App.~\ref{app:human_pilot}). The full generator, the 3{,}931-question benchmark, and the evaluation harness are released publicly, together with a Python API for the generator and evaluation harness.\footnote{\url{https://github.com/TiandaSun/Kinship-Data-Benchmark-for-Multi-hop-Reasoning}; CC BY 4.0.}

\subsection{Question Categories}

We define four question categories of progressively more complex reasoning. \textbf{\mbox{Cat.~1 (Fact Retrieval)}} is a 1-hop lookup of an explicitly stated relation (\emph{Who is John's father?}), a baseline for context extraction. \textbf{Cat.~2 (Multi-Hop Biological)} chains 2--4 biological steps (\emph{paternal grandfather} = father$\to$father), testing compositional reasoning over universal genealogical structure. \textbf{Cat.~3 (Counting/Filtering)} adds logical operations on top of Cat.~2: counting (\emph{How many siblings does Mary have?}), filtering (\emph{List X's male cousins}), and comparison. \textbf{Cat.~4 (Cultural Disambiguation)} tests whether LLMs can deploy culture-specific classification rules that diverge from the biological default. The model receives the system name and an in-context cultural query (\emph{According to Hawaiian kinship, who are X's parents?} relies on the convention that mother's sisters are classified as ``mothers''). \emph{Explicit rule statements are never given in the prompt}: Cat.~4 therefore tests application of a \emph{named latent rule system}---combining recall of the cultural prior with override behaviour---rather than application of an explicitly supplied in-context rule. The fictional-rule control (\S\ref{sec:interventions}) probes the contribution of cultural-prior interference to the residual failure.

\begin{table}[t]
\centering
\caption{KinshipQA dataset statistics for the 1--4 hop subset (3{,}134 questions); 5--6 hop Cat.~2 coverage adds 548 further questions across all 7 systems (App.~\ref{app:hop_models}). ``Override'' is the fraction of questions requiring cultural reclassification (e.g.\ mother's sister $\rightarrow$ classificatory mother in Hawaiian); only Cat.4 questions on the Other-5 systems are override-eligible. Sudanese is 0\% because its relational structure aligns with the biological default---its lexical uniqueness (a distinct term per kin relation) does not constitute structural override.}
\small
\begin{tabular}{lrrr}
\toprule
& \textbf{Questions} & \textbf{\%} & \textbf{Override} \\
\midrule
\multicolumn{4}{l}{\textit{By System Type}} \\
\quad Eskimo, Sudanese & 974 & 31.1 & 0\% \\
\quad Remaining five & 2,160 & 68.9 & 30.6\% \\
\midrule
\multicolumn{4}{l}{\textit{By Category}} \\
\quad Cat 1: Fact & 350 & 11.2 & -- \\
\quad Cat 2: Multi-hop & 1,050 & 33.5 & -- \\
\quad Cat 3: Order \& count & 700 & 22.3 & -- \\
\quad Cat 4: Cultural & 1,034 & 33.0 & 63.8\% \\
\midrule
\multicolumn{4}{l}{\textit{By Complexity}} \\
\quad 1-hop & 985 & 31.4 & -- \\
\quad 2-hop & 877 & 28.0 & -- \\
\quad 3-hop & 812 & 25.9 & -- \\
\quad 4-hop & 460 & 14.7 & -- \\
\midrule
\textbf{Total} & \textbf{3,134} & \textbf{100} & \textbf{21.1\%} \\
\bottomrule
\end{tabular}
\label{tab:stats}
\end{table}
\subsection{Comparing LLMs with KinshipQA}
\label{sec:datasets}
We have used our pipeline to generate one dataset for each type of kinship system. All datasets span 50 years, starting from a seed population of four individuals and growing to 60--70 individuals under the hard-wired offspring constraints (full simulator parameters in the supplementary repository). The resulting trees are used to evaluate and compare a number of LLMs popular at present.

Table~\ref{tab:stats} reports overall statistics; per-cell counts (system\,$\times$\,category\,$\times$\,hop) are released alongside the dataset. The largest cells (e.g.\ Cat.~2 at 2--3 hops on each system) contain 100--350 questions. The cultural-override path library covers 2-, 3-, and 4-hop paths per Other-5 system, anthropologically grounded in Morgan's typology (e.g., Hawaiian: parallel-uncle's grandchild $\to$ classificatory child; Crow: father's sister's grandchild $\to$ skewed matrilineage member; full path table in Appendix~\ref{app:cat4_4hop}). At 4-hop, this contributes 34--50 Cat.~4 questions per Other-5 system (Hawaiian/Iroquois/Dravidian: 50; Crow: 36; Omaha: 34; 220 in total)---sufficient for per-cell mean estimates with bootstrap CIs as reported in \S\ref{sec:hop_matched}.

\paragraph{Generation controls.} Templates are organised by category and hop depth (4 / 12 / 6 / per-system override templates for Cat.~1--4 respectively); names are sampled uniformly from a fixed roster of common first names with 4 surnames per tree; and context paragraphs include only the minimal subgraph required for the SPARQL ground-truth derivation, with no added distractor facts---reasoning difficulty derives from chain length and cultural-rule application rather than information-filtering noise.


\section{Results and Evaluation}
\subsection{Experimental Setup}
\label{sec:setup}

\paragraph{Models and protocols.} We evaluate six LLMs: three open-source 27--32B models --- Qwen3-32B \citep{yang_qwen3_2025}, Gemma3-27B \citep{team_gemma_2025}, DeepSeek-R1-Distill-Qwen-32B (hereafter DeepSeek-R1-32B; \citealp{deepseek-ai_deepseek-r1_2025}) --- and three closed-source mid-tier APIs --- GPT-4o-mini \citep{openai_gpt-4o_2024}, Claude-Haiku-4.5 \citep{anthropic_claude_2024}, Gemini-2.5-Flash \citep{comanici_gemini_2025}. All models receive identical zero-shot prompts under greedy decoding (temperature\,$=0$). We additionally evaluate the open-source models and two closed-source models (GPT-4o-mini, Claude-Haiku-4.5) under zero-shot and few-shot CoT; Gemini-2.5-Flash is omitted from the CoT matrix due to free-tier quota.

\paragraph{Metric and infrastructure.} We report \textbf{Exact Match (EM)} as our primary metric; multiple-valid-answer questions (43.8\%) are scored set-based.\footnote{Bootstrap CIs use $N{=}10{,}000$ percentile-method resamples (unit: question for hop-matched comparisons, (system, model) cell for ablations; unpaired, stratified within group).} Because the symbolic SPARQL executor returns the gold on 100\% of questions by construction (\S\ref{sec:datasets}), the LLM gaps reported below isolate natural-language interpretation and reasoning rather than evaluator drift. Open-source models run via Ollama on a single H100 80GB GPU (Q4\_K\_M quantization, \textit{max\_tokens}\,$=512$ for direct / $2048$ for CoT); closed-source models use vendor APIs with matched settings. Model identifiers, parsers, prompt templates, and the dataset generator are released in the public repository (\S\ref{sec:datasets}).

\begin{table*}[t]
\centering
\caption{Main results on KinshipQA (Exact Match \%; F1 in brackets for the multi-answer categories Cat.~3 and Cat.~4). 
\textbf{Scope:} Cat.~1--3 are computed on all 7 systems; Cat.~4 is computed only on the Other-5 systems (Hawaiian, Iroquois, Dravidian, Crow, Omaha), since Esk\&Sud (Eskimo+Sudanese) are descriptive controls without override paths by construction. $\Delta$ Gap $=$ Other-5 $-$ Esk\&Sud over Cat.~1--4 (negative entries indicate a drop on the Other-5 systems). All six models are evaluated under zero-shot direct prompting with greedy decoding (temperature\,$=0$); the Avg.\ row is the unweighted mean over the six models. Under F1 (partial-credit set match), Cat.~4 Other-5 scores rise by 8--11\%, but the cultural-override gap persists: the F1 gap remains +18 to +43\% across models, preserving the same descriptive\,$\to$\,skewing hierarchy.}
\small
\begin{tabular}{lcccccccc}
\toprule
& \multicolumn{3}{c}{\textbf{By Culture}} & & \multicolumn{4}{c}{\textbf{By Category}} \\
\cmidrule{2-4} \cmidrule{6-9}
\textbf{Model} & Esk\&Sud & Other 5 & $\Delta$ Gap & & Cat.1 & Cat.2 & Cat.3 (F1) & Cat.4 (F1) \\
\midrule
\multicolumn{9}{l}{\textit{Closed-Source Models}} \\
GPT-4o-mini & 92.3 & 81.2 & $-$11.1 & & 100.0 & 90.4 & 86.1 (86.1) & 62.6 (72.5) \\
Claude-Haiku-4.5 & 98.3 & 85.3 & $-$13.0 & & 100.0 & 98.3 & 87.7 (87.9) & 65.0 (72.8) \\
Gemini-2.5-Flash & 98.2 & 84.3 & $-$13.9 & & 100.0 & 99.5 & 86.1 (86.1) & 61.4 (70.8) \\
\midrule
\multicolumn{9}{l}{\textit{Open-Source Models}} \\
Qwen3-32B & 92.2 & 77.9 & $-$14.3 & & 99.4 & 96.0 & 86.1 (86.1) & 44.8 (53.2) \\
Gemma3-27B & 96.8 & 84.7 & $-$12.1 & & 100.0 & 98.4 & 84.9 (85.7) & 65.3 (73.4) \\
DeepSeek-R1-32B & 92.7 & 73.8 & $-$18.9 & & 89.1 & 94.6 & 82.1 (88.1) & 41.1 (52.0) \\
\midrule
\textbf{Avg.} & 95.1 & 81.2 & $-$13.9 & & 98.1 & 96.2 & 85.5 (86.7) & 56.7 (65.8) \\
\bottomrule
\end{tabular}
\label{tab:main_results}
\end{table*}

Table~\ref{tab:main_results} presents our main findings under zero-shot direct prompting. Table~\ref{tab:protocol_comparison} extends to all three protocols (zero-shot direct, zero-shot CoT, few-shot CoT) on the five models with closed-source coverage stable enough to run the full matrix; Gemini-2.5-Flash is omitted from CoT protocols due to API quota.

\begin{table}[t]
\centering
\caption{Accuracy by protocol on the five models with full protocol-matrix coverage (Exact Match \%). \textbf{Overall} averages all four categories on all 7 systems; \textbf{Cat.~4} reports Other-5 Cat.~4 EM. Few-shot CoT raises Overall on 4 of 5 models and Cat.~4 Other-5 on 3 of 5; the cultural-override gap persists in every cell. Per-system protocol breakdown in App.~\ref{app:by_system_protocol}.}
\label{tab:protocol_comparison}
\small
\setlength{\tabcolsep}{4pt}
\begin{tabular}{l ccc ccc}
\toprule
& \multicolumn{3}{c}{\textbf{Overall (\%)}} & \multicolumn{3}{c}{\textbf{Cat.4 (\%)}} \\
\cmidrule(lr){2-4} \cmidrule(lr){5-7}
\textbf{Model} & Dir. & CoT & FS & Dir. & CoT & FS \\
\midrule
GPT-4o-mini & 84.4 & \textbf{89.1} & 87.5 & 62.6 & \textbf{65.0} & 56.1 \\
Claude-Haiku-4.5 & 89.0 & 89.3 & \textbf{89.6} & 65.0 & 65.0 & \textbf{67.0} \\
Gemma3-27B & 88.2 & 60.7 & \textbf{88.9} & 65.3 & 37.0 & 63.6 \\
DeepSeek-R1 & 79.2 & 74.1 & \textbf{86.9} & 41.1 & 39.2 & \textbf{52.9} \\
Qwen3-32B & 82.0 & 83.8 & \textbf{87.8} & 44.8 & 49.8 & \textbf{57.6} \\
\midrule
\textbf{Avg.} & 84.6 & 79.4 & \textbf{88.1} & 55.8 & 51.2 & \textbf{59.4} \\
\bottomrule
\end{tabular}
\end{table}

Few-shot CoT is best on overall accuracy for 4 of 5 models (Avg.\ $84.6 \to 88.1$\%) and raises Cat.~4 Other-5 accuracy by $+3.6$\% on average. Gemma3 and GPT-4o-mini drop slightly on Cat.~4 under FS-CoT (Gemma3 $65.3 \to 63.6$\%, GPT-4o-mini $62.6 \to 56.1$\%) despite gains on Cat.~1--3, consistent with demonstrations drawn from other systems biasing these models toward the demonstrated systems' rules. In every protocol on every model, the Esk\&Sud descriptive controls reach 92--100\% Cat.~4 accuracy while the Other-5 systems plateau at 37--67\% (\S\ref{sec:prompt_robust}).

\subsection{Performance by Reasoning Complexity}
\label{sec:hops}

\begin{figure}[t]
\centering
\includegraphics[width=\columnwidth]{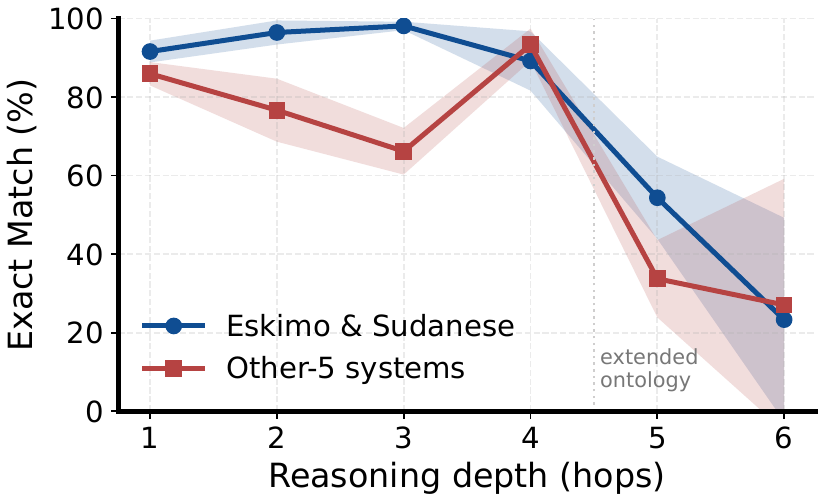}
\caption{Overall performance by reasoning depth (EM \%, zero-shot direct, mean across the three open-source models; shaded bands: $\pm$ cross-model std), split into Esk\&Sud (descriptive controls) and the Other-5 systems where cultural override applies. The 1--4 hop average mixes Cat.~1--4; the 5--6 hop average covers 548 Cat.~2 questions on all 7 systems plus 249 Cat.~4 questions on the Other-5 systems (see \S\ref{sec:hops} and Appendix~\ref{app:cat4_56hop}). The vertical dotted line at 4.5 marks the simulator boundary between the 50-year ontology and the extended 150/200-year ontologies. The 5--6 hop points use the 8{,}192-token re-run of all empty predictions (\S\ref{sec:hops}); the original 512-token sweep truncated the reasoning models before the answer line.}
\label{fig:nhops}
\end{figure}

To probe scaling beyond four hops, we regenerate ontologies with extended genealogies for all 7 systems (150 years for Eskimo, Sudanese, Iroquois, Dravidian; 200 years for Hawaiian, Crow, Omaha, whose more restrictive marriage rules require a longer horizon). This yields 548 new 5--6 hop Cat.~2 questions across all 7 systems and 249 5--6 hop Cat.~4 cultural-override questions on the Other-5 systems (Esk\&Sud have no override paths by construction). Per-hop Cat.~4 accuracy on the 50-year ontology is visualised in App.~\ref{app:hop_models}, Figure~\ref{fig:cat4_hop}.

Within Cat.~4 at 1--4 hops, accuracy on the Other-5 systems drops monotonically (58.4\,$\to$\,42.5\,$\to$\,18.9\%) while Esk\&Sud declines only modestly (96.0\,$\to$\,94.4\,$\to$\,85.5\%); the cultural-override gap widens from 37.6\% (2-hop) to 66.5\% (4-hop), suggesting that cultural disambiguation and chain depth compound. At 5--6 hops, Other-5 Cat.~4 EM falls further to 10.6\% ($-$39.9\% from the 1--4 hop baseline of 50.4\%; per-system drops range $-7$ to $-52$\%, App.~\ref{app:cat4_56hop}). The original 5--6 hop sweeps ran at a 512-token generation budget that truncated the reasoning models before the answer line; re-running every empty prediction at 8{,}192 tokens raises Cat.~2 accuracy from 46.4\% to 62.7\% at five hops and from 26.3\% to 52.3\% at six (App.~\ref{app:hop_models}). Biological composition therefore degrades gracefully with depth while cultural override does not: on the same Other-5 chains at 5--6 hops, Cat.~2 reaches 58.6\% against 10.6\% for Cat.~4. The within-Other-5 ordering partially scrambles at this depth (Omaha 25.3\%, Crow 0.0\% on a small 16-question cell with multi-name targets; we read the Crow EM against its F1 of 7.9\% rather than in isolation), reflecting per-system path-structure differences at the extended-horizon ontology boundary. On the overall 1--4 hop curve (Figure~\ref{fig:nhops}), apparent local non-monotonicities are category-composition shifts rather than depth reversals, since Cat.~4 tapers from the Other-5 average as depth grows; full per-model curves in App.~\ref{app:hop_models}, consistent with the compositionality limits of \citet{dziri_faith_2023} and \citet{lin_zebralogic_2025}.

\subsection{Performance by Kinship System}

\begin{table}[t]
\centering
\caption{Performance by kinship system (EM \%, mean $\pm$ std across 3 open-source models; ordered descriptive first, skewing last). ``All Cat.''\ averages Cat.~1--4. The right block reports Cat.~4 with cultural override (applicable only to Other-5 systems; Esk\&Sud entries ``--'') against the per-system mean over Cat.~1--3. Humans given the same override rule reach 89.0\% on Other-5 Cat.~4 vs.\ 50.7\% for LLMs (App.~\ref{app:human_pilot}). The descriptive-versus-skewing contrast holds under F1 (per-system Cat.~4 F1 in App.~\ref{app:rule_context}).}
\small
\setlength{\tabcolsep}{4pt}
\begin{tabular}{lcccc}
\toprule
\textbf{System} & \textbf{All Cat.} & \textbf{w/ override} & \textbf{Cat.~1--3} & \textbf{$\Delta$} \\
\midrule
Eskimo    & 94.1 $\pm$ 2.1 & --              & 94.8 $\pm$ 1.9 & --   \\
Sudanese  & 93.7 $\pm$ 3.0 & --              & 94.8 $\pm$ 1.9 & --   \\
\midrule
Hawaiian  & 82.6 $\pm$ 3.0 & 61.9 $\pm$ \phantom{0}7.2 & 91.8 $\pm$ 2.7 & 29.9 \\
Iroquois  & 78.2 $\pm$ 5.5 & 50.3 $\pm$ 14.8 & 90.6 $\pm$ 1.6 & 40.3 \\
Dravidian & 81.2 $\pm$ 6.8 & 60.4 $\pm$ 16.6 & 90.3 $\pm$ 3.3 & 30.0 \\
Crow      & 78.4 $\pm$ 6.7 & 47.0 $\pm$ 15.7 & 92.2 $\pm$ 3.1 & 45.3 \\
Omaha     & 73.7 $\pm$ 5.8 & 32.6 $\pm$ 13.7 & 91.8 $\pm$ 2.7 & \textbf{59.2} \\
\midrule
\multicolumn{2}{l}{\textit{Other-5 avg.}} & 50.4 $\pm$ 11.8 & 91.3 $\pm$ 0.8 & 40.9 \\
\bottomrule
\end{tabular}
\label{tab:by_system}
\end{table}

Table~\ref{tab:by_system} reveals a clear separation across system types: Eskimo and Sudanese sit near ceiling, the bifurcate (Iroquois, Dravidian) and generational (Hawaiian) systems are intermediate, and the two skewing systems (Crow, Omaha) are hardest---Omaha bottoming at 32.6\% Cat.~4 EM (45.6\% F1). We take the robust pattern to be the descriptive-versus-non-descriptive contrast together with the skewing penalty; the ordering \emph{among} Hawaiian, Iroquois and Dravidian is not stable across metrics and depths (App.~\ref{app:cat4_56hop}) and we do not claim it. Comparing per-system Cat.~4 (cultural-rule application) against Cat.~1--3 (biological reasoning) highlights the empirical cost of the override setting: switching from biological chaining to cultural-rule application drops accuracy by 40.9\% EM (and 31.8\% F1) on average across the Other-5 systems, with the largest gaps in skewing systems (Omaha 59.2\% EM, Crow 45.3\% EM) and the smallest in Hawaiian and Dravidian (29.9 and 30.0\% EM); under F1 the same descriptive-versus-skewing contrast holds (Omaha 45.6, Crow 53.2, Iroquois 63.2, Dravidian 65.9, Hawaiian 69.8). Generational merging (Hawaiian) and cross-cousin rules (Dravidian) thus appear more learnable than generational skewing on both metrics. Human annotators (App.~\ref{app:human_pilot}; $n=280$, two graduate-student volunteers, IAA 96.8\%) reach 89.0\% on Other-5 Cat.~4 when given the same one-sentence rule as the in-context-rule probe, compared with 50.7\% for LLMs under the same rule-supplied setting. This indicates the questions are intelligible under matched rule access rather than ambiguous.

Each step up the hierarchy adds a structural demand: descriptive systems impose no override, generational and bifurcate require a one-generation reclassification (sex-and-generation matching; parallel-vs-cross), and skewing requires generation-spanning reclassification keyed to lineage membership --- the most counterintuitive override and the rarest in pretraining corpora, consistent with the in-context-rule probe's evidence for a weaker skewing prior (\S\ref{sec:interventions}).

\section{Cultural Findings}
\label{sec:cultural}

\subsection{Robustness: Not a Prompting or Depth Artefact}
\label{sec:robustness}

\paragraph{Prompting.}\label{sec:prompt_robust} The Esk\&Sud~$\to$~Other-5 gap \textbf{persists under all three protocols} for the five models with full protocol-matrix data (App.~\ref{app:gap_by_protocol}). Few-shot CoT raises Cat.~4 Other-5 accuracy by $+3.6$\% on average but shifts the gap inconsistently across models. In every protocol on every model, the descriptive controls reach 92--100\% Cat.~4 accuracy while the Other-5 systems plateau at 37--67\%, consistent with \citet{sprague_cot_2025}'s finding that CoT helps mainly on symbolic-format reasoning.

\paragraph{Chain depth.}\label{sec:hop_matched}

\begin{table}[t]
\centering
\caption{Hop-matched comparison of Cat.2 (biological multi-hop, all 7 systems) vs Cat.4 (cultural override, Other-5 systems) on identical input format. The gap \textbf{widens} from 38.0\% at 2-hop to 56.3\% at 3-hop, opposite to a lookup-disruption prediction and suggesting that cultural-rule application adds difficulty beyond chain depth. Per-model breakdown shows the gap holds across all three open-source models. (DeepSeek-R1-32B's small Cat.2 increase from 2-hop (89.7) to 3-hop (95.4) reflects per-cell sample variance; the cross-model average and the cultural-gap pattern are unchanged by it.)}
\label{tab:hop_matched}
\small
\begin{tabular}{llccc}
\toprule
\textbf{Model} & \textbf{Hop} & \textbf{Cat.2} & \textbf{Cat.4} & \textbf{Gap} \\
\midrule
Gemma3-27B & 2 & 100.0 & 76.7 & 23.3 \\
 & 3 & 100.0 & 53.9 & 46.1 \\
\midrule
DeepSeek-R1-32B & 2 & 89.7 & 48.8 & 40.9 \\
 & 3 & 95.4 & 33.3 & \textbf{62.1} \\
\midrule
Qwen3-32B & 2 & 100.0 & 50.3 & 49.7 \\
 & 3 & 100.0 & 39.4 & 60.6 \\
\midrule
\textbf{Average} & \textbf{2} & \textbf{96.6} & \textbf{58.6} & \textbf{38.0} \\
\textbf{Average} & \textbf{3} & \textbf{98.5} & \textbf{42.2} & \textbf{56.3} \\
\bottomrule
\end{tabular}
\end{table}

Could the Cat.2$\to$Cat.4 drop simply reflect task difficulty rather than cultural variation? Table~\ref{tab:hop_matched} compares Cat.2 and Cat.4 \textbf{at matched hop depths}. The cultural gap is substantial at every cell (23.3--62.1\%) and \emph{widens} with depth (38.0\% at 2-hop $\to$ 56.3\% at 3-hop), opposite to a lookup-disruption prediction. These results suggest that cultural override and multi-hop complexity compound.

\paragraph{Phrasing, decoding, and frontier scale.}\label{sec:phrasing_robust} A rule-paraphrased Cat.~4 subsample on Gemma3-27B preserves accuracy (78.0\,$\to$\,84.0\%; App.~\ref{app:paraphrase_robust}), and multi-seed decoding ($T{=}0.7$, $n{=}5$) is 3--14\% lower than greedy but preserves the system-type ordering (App.~\ref{app:multi_seed}). At frontier scale, Claude Opus 4.7 reaches 100\% on Esk\&Sud but 69\% Other-5 under zero-shot direct (gap 31\%, narrowing to 21.6\% under FS-CoT, App.~\ref{app:opus_check}); the depth\,$\times$\,override interaction also widens at frontier scale (22.4\,$\to$\,47.6\% from 2- to 3-hop).

\paragraph{Output format.}\label{sec:format_robust} Unparseable outputs do not drive the gap. On the extended-horizon sweep, where the empty/format errors concentrate, the Cat.~1--3 vs.\ Cat.~4 gap is 33.7\% as scored, 31.0\% after dropping every unparseable prediction from both sides, and 27.7\% after re-running all 1{,}016 empty predictions at 8{,}192 tokens. Those empties are generation-length truncations rather than refusals: 89.7\% return an answer at the larger budget, and 64.9\% of the formerly-empty Cat.~4 items are then correct. Biological-default substitution remains the largest substantive error class under every one of the three treatments.

\subsection{Probing Cat.~4 Failure: Two Interventions}
\label{sec:interventions}

Cat.~4 errors are dominated by biological-default leakage. Of the 1{,}000 non-EM predictions, 358 return no parseable answer; among the 642 \emph{substantive} errors, 83.0\% are bare-default substitution (51.1\%) or over-inclusion (31.9\%), and only 15.3\% are wrong-disjoint (App.~\ref{app:cat4_taxonomy}). Two interventions probe this gap.

\begin{table}[t]
\centering
\small
\setlength{\tabcolsep}{3pt}
\caption{Fictional-rule control: orthogonal decomposition of the $+6.1$\% Cat.~4 gain into system-name and kin-term swaps (Other-5 systems, mean across 3 open-source models, $\Delta$ vs.\ real-label baseline; 95\% bootstrap CI over 15 (system, model) cells). Either swap alone is null; the joint swap is $+6.1$, indicating an interaction.}
\label{tab:fictional_orthogonal}
\begin{tabular}{lccc}
\toprule
Condition & Real EM & $\Delta$ vs. Real & 95\% CI \\
\midrule
Real (baseline)      & 50.4 & -- & -- \\
+ system-name swap   & 51.7 & +1.3 & [$-$1.2, +4.1] \\
+ kin-term swap      & 49.2 & $-$1.2 & [$-$4.4, +2.3] \\
+ both (full)        & 56.5 & \textbf{+6.1} & \textbf{[+2.8, +9.4]} \\
\bottomrule
\end{tabular}
\end{table}

\begin{table}[t]
\centering
\small
\setlength{\tabcolsep}{4pt}
\caption{In-context-rule probe: mean $\Delta$EM (\%) by override typology and model baseline ($n$ = model$\times$system pairs). Skewing cells uniformly help; non-skewing high-baseline cells uniformly hurt. Full per-cell data in App.~\ref{app:rule_context}.}
\label{tab:rule_2x2}
\begin{tabular}{lcc}
\toprule
\textbf{Baseline EM} & \textbf{Non-skewing} & \textbf{Skewing} \\
\midrule
$\geq$~60\%   & $-$13.4 ($n{=}4$) & $+$12.0 ($n{=}1$) \\
$<$~60\%      & $+$4.9 ($n{=}5$) & $+$17.9 ($n{=}5$) \\
\bottomrule
\end{tabular}
\end{table}

\paragraph{Fictional-rule control.} Replacing both the system label and the kin terms with phonotactically distinct invented strings raises Cat.~4 Other-5 accuracy by $+6.1$\% (95\% CI [+2.8, +9.4]); replacing either alone is null (Table~\ref{tab:fictional_orthogonal}), so the gain is a surface-feature \emph{interaction} rather than additive. A pure knowledge deficit predicts the opposite direction; accuracy instead rises on all three open-source models (per-model Other-5 means: Gemma3 $+1.1$, DeepSeek $+5.7$, Qwen3 $+11.5$), strongest on the hardest systems (Iroquois, Dravidian, Omaha) and the two reasoning-focused models, pointing to familiar surface forms as one contributor to the Cat.~4 gap (per-cell breakdown in App.~\ref{app:fictional_control}). We read this as \emph{consistent with} surface-form interference rather than as isolating it: altered answer-space expectations, reduced prompt ambiguity, and differential parsing around culturally loaded words are not excluded by this control.

\paragraph{In-context rule probe.} If failure were instead a deficit of rule \emph{recall}, supplying the rule in context should close the gap. We prepend a one-sentence override rule to each Cat.~4 question and evaluate the three open-source models on the full Other-5 set (1{,}254 questions/model; App.~\ref{app:rule_context}). The effect splits along a typology\,$\times$\,baseline axis (Table~\ref{tab:rule_2x2}). Skewing systems show a uniform rule-recall deficit (6/6 pairs HELP, mean $+17.1$\%; per-model on Crow/Omaha: Gemma3 $+12/+28$, DeepSeek $+14/+19$, Qwen3 $+6/+23$). Non-skewing systems above $\sim$60\% baseline EM are \emph{hurt} by the rule ($-9$ to $-18$\%; 4/4 pairs, mean $-13.4$), consistent with interference between the supplied rule and an internalised approximation; below $\sim$60\% the rule helps or is flat (descriptive-control noise floor $\leq 7.5$\%).

\section{Conclusion}

KinshipQA spans seven anthropologically-documented kinship systems with a cultural-disambiguation category. The 40.9\% Cat.~1--3\,$\to$\,Cat.~4 drop holds for every one of the five non-descriptive systems and is largest for the two skewing systems (Table~\ref{tab:by_system}), persists under chain-of-thought, few-shot, stochastic decoding, depth-matched controls, and frontier-scale evaluation, and compounds with chain depth: at 5--6 hops cultural override falls to 10.6\% while biological composition over the same chains holds at 58.6\%. Two interventions help distinguish two contributing factors: removing familiar surface forms raises accuracy by 6.1\% (pointing to surface forms as one contributor); supplying the override rule helps skewing systems but reduces accuracy on non-skewing systems with high baseline (consistent with a knowledge gap and rule interference respectively). Humans given the same rule reach 89.0\% vs.\ 50.7\% for LLMs --- a 38.3\% gap, so the questions are reliably solvable once the rule is supplied. Benchmarks limited to a single descriptive typology may therefore overestimate how reliably models apply kinship rules that override the English default.

\section*{Limitations}

KinshipQA covers seven kinship systems from Morgan's typology, implemented as idealised anthropological models in English; the named systems are analytical abstractions, not rigid descriptions of contemporary practice, and the simulator assumes a binary-sex two-parent biological family structure. Our ``cultural disambiguation'' construct captures performance on stylised English glosses of Morgan-typology overrides; pretraining-frequency effects of these specific glosses are not separately measured. Contexts contain only the minimal subgraph required for SPARQL ground truth (no distractor facts), so KinshipQA does not test evidence selection or robustness to irrelevant facts. Kinship semantics are specified through a predefined symbolic ontology rather than induced from data. Closed-source coverage in our headline tables is the cost-controlled mid-tier (GPT-4o-mini, Claude-Haiku-4.5, Gemini-2.5-Flash); we evaluate Claude Opus 4.7 at frontier scale on a Cat.~4 subset (Appendix~\ref{app:opus_check}). Fine-tuning and retrieval-augmented regimes are out of scope. Symbolic gold answers are exact by construction; a two-annotator human baseline ($n=280$, IAA 96.8\%; App.~\ref{app:human_pilot}) reaches 95.4\% overall and 89.0\% on Other-5 Cat.~4 with the same in-context rule used in the in-context-rule probe (\S\ref{sec:interventions}); the like-for-like LLM-with-rule mean is 50.7\%. Multilingual native-speaker validation is left to future work.

\section*{Acknowledgments}

Generative AI tools (Claude, Anthropic) were used during revision for language polishing and code review; all research contributions and substantive analyses were authored and verified by the human authors.

\bibliography{custom-Dimitar}

\appendix

\section{Kinship Systems and Override Mappings}
\label{app:kinship-systems}

\subsection{System Descriptions}

\paragraph{Eskimo (Lineal).} Distinguishes lineal relatives (parents, grandparents) from collateral relatives (aunts, uncles, cousins). Mother's sister and father's sister both = ``aunt.''

\paragraph{Hawaiian (Generational).} Merges all relatives of the same generation and sex. All female relatives of mother's generation = ``mother''; all male relatives of father's generation = ``father''; Cousins = siblings.

\paragraph{Iroquois (Bifurcate Merging).} Distinguishes parallel relatives (same-sex parents' sibling) from cross relatives (opposite-sex parents' sibling). Father's brother = ``father''; mother's brother = ``uncle.'' Parallel cousins = siblings; cross cousins distinguished.

\paragraph{Dravidian (Bifurcate with Cross-Cousin Marriage).} Similar to Iroquois, but cross-cousins are prescribed marriage partners, classified as potential spouses rather than kin.

\paragraph{Crow (Matrilineal Skewing).} Members of father's matrilineage are ``skewed'' upward generationally. Father's sister's children are classified as ``fathers/female fathers'' regardless of actual generation.

\paragraph{Omaha (Patrilineal Skewing).} Mirror image of Crow. Members of the mother's patrilineage are skewed upward. Mother's brother's children are classified as ``mothers/male mothers.''

\paragraph{Sudanese (Descriptive).} Maximally descriptive system with unique terms for each biological relationship. No merging or classification rules—each kin type receives a distinct label.

\subsection{Cultural Override Mappings}
\label{app:mappings}

Table~\ref{tab:overrides} shows how biological relationship paths map to different kinship terms across cultural systems.

\begin{table}[h]
\centering
\setlength{\tabcolsep}{3pt}
\small
\caption{Cultural override mappings for key relationship paths. Note that cl. = classificatory}
\begin{tabular}{llll}
\toprule
\textbf{Bio Path} & \textbf{Esk./Sud.} & \textbf{Hawaiian} & \textbf{Iroq./Drav.} \\
\midrule
F→Brother & pat. uncle & cl. father & cl. father \\
M→Sister & mat. aunt & cl. mother & cl. mother \\
F→Sister & pat. aunt & cl. mother & cross-aunt \\
M→Brother & mat. uncle & cl. father & cross-uncle \\
F→Bro→Child & cousin & cl. sibling & parallel cousin \\
M→Sis→Child & cousin & cl. sibling & parallel cousin \\
F→Sis→Child & cousin & cl. sibling & cross-cousin \\
M→Bro→Child & cousin & cl. sibling & cross-cousin \\
\bottomrule
\end{tabular}
\label{tab:overrides}
\end{table}

\section{Qualitative Examples}
\label{app:qualitative examples}

\paragraph{Example 1: Chain Tracking Failure (Eskimo System)}

\begin{quote}
\textbf{Question:} Who are Lisa Williams's paternal uncle's grandchildren?

\textbf{Context:} Lisa's father is Robert Williams. Robert's brother is Larry Williams. Larry's children are Patricia, Justin, and Mark. Mark's children are Samantha, Jerry, and Jason. Justin's children are Betty, Nicholas, and Ashley.

\textbf{Ground Truth:} 6 people (Samantha, Jerry, Jason, Betty, Nicholas, Ashley)

\textbf{Prediction:} Patricia, Justin, Mark (3 people)
\end{quote}

\textbf{Analysis:} The model returns Larry's \textit{children} (3 hops) instead of \textit{grandchildren} (4 hops). This error occurs in the Eskimo (descriptive lineal) system, demonstrating that chain complexity---not cultural-rule application---is the limiting factor.

Example 1 illustrates a chain-tracking failure in the descriptive (Eskimo) system; the cultural-default and generational-skewing failure modes are quantified across all 1{,}000 non-EM Cat.~4 predictions in the automatic error taxonomy of \S\ref{app:cat4_taxonomy}.

\section{Prompt Templates and 4-hop Cultural-Override Paths}
\label{app:prompts}

\subsection{Zero-Shot Evaluation Prompt}
\begin{small}
\begin{verbatim}
Answer the following question based
on the given context. 
Be concise and provide only
the answer without explanation.

Context: {context}

Question: {question}

Answer:
\end{verbatim}
\end{small}

\subsection{Zero-Shot Chain-of-Thought Prompt}
\begin{small}
\begin{verbatim}
You are analyzing kinship relationships in a family.
Read the context carefully and answer the question
by showing your complete reasoning process.

Context: {context}

Question: {question}

Please think through this step-by-step:

STEP 1 - IDENTIFY KEY PERSON(S):
Who is the question asking about?
What relationship are we looking for?

STEP 2 - EXTRACT RELEVANT FACTS:
List the specific family relationships from the
context that are relevant.

STEP 3 - TRACE THE REASONING CHAIN:
Work through the relationships step by step.
Show each connection.

STEP 4 - APPLY CULTURAL RULES (if mentioned):
If the context mentions any specific kinship system
rules (like clan membership, classificatory
relationships, or cultural terminology), apply them
here.

STEP 5 - FINAL ANSWER:
End with the line "FINAL ANSWER: <answer>".
\end{verbatim}
\end{small}

\subsection{Few-Shot CoT Prompt (sketch)}
For few-shot CoT we prepend three worked examples per question category to the zero-shot CoT template above. Each worked example contains a short context, a question, the five reasoning steps, and a \texttt{FINAL ANSWER} line, illustrating the expected reasoning format and an instance of cultural-rule application where applicable.

\subsection{4-hop Cultural-Override Paths}
\label{app:cat4_4hop}

The cultural-override path library includes anthropologically grounded paths at hop depths 2, 3, and 4 per Other-5 system, supporting within-Cat.4 hop-scaling analysis in \S\ref{sec:hops}. At 4-hop, the library produces $\sim$50 Cat.4 questions per Other-5 system. The per-system path mappings extend the same logic as the 2-hop overrides documented in Table~\ref{tab:overrides} (App.~\ref{app:mappings}): same-sex chains remain parallel in Iroquois/Dravidian; FZ-line / MB-line members remain skewed across generations in Crow/Omaha; G$\pm$1 relatives become own parents/children in Hawaiian. Full mappings are in the supplementary repository.

\begin{figure}[h]
\centering
\includegraphics[width=0.78\columnwidth]{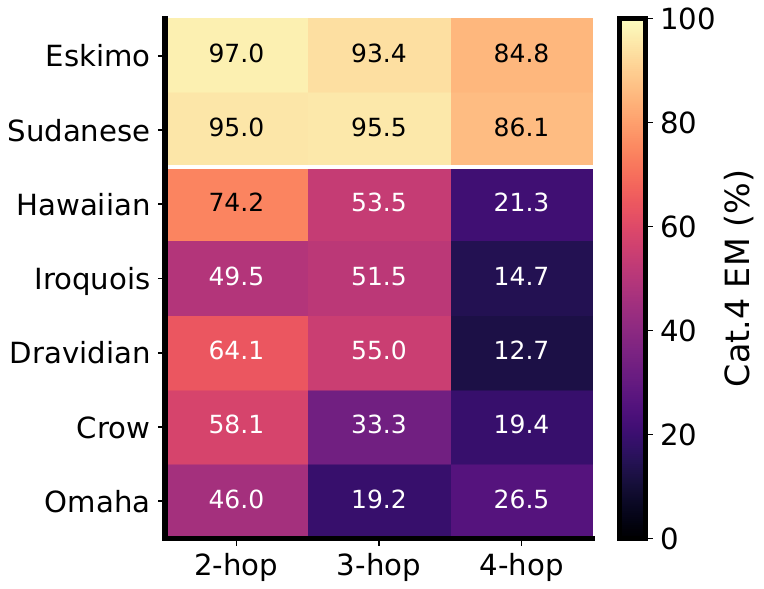}
\caption{Per-system Cat.4 EM (\%) by hop (zero-shot direct, mean across the three open-source models). Eskimo and Sudanese (above the white divider) serve as descriptive controls; the five Other-5 systems below decline sharply with depth, with the bifurcate systems (Dravidian, Iroquois) and Omaha hitting the lowest 4-hop accuracies.}
\label{fig:cat4_per_system_hop}
\end{figure}

\paragraph{Cross-model variance at 4-hop.} The cross-model std rises sharply at 4-hop (e.g., Hawaiian $\pm$24.6, Omaha $\pm$29.2; visualised per-system in Figure~\ref{fig:cat4_per_system_hop}). Per-model breakdowns show a stark split: Gemma3-27B retains 34--67\% accuracy on 4-hop Other-5 Cat.4 while DeepSeek-R1-32B and Qwen3-32B collapse to 0--11\%. The two reasoning-focused models are more affected by the depth\,$\times$\,override interaction than the general-purpose Gemma3, an observation we leave to future work.

\newpage

\section{Error Analysis Detail}
\label{app:errors}

\paragraph{Automatic Cat.~4 error decomposition.}\label{app:cat4_taxonomy} We automatically classify all 1{,}000 non-EM Cat.~4 predictions across the three open-source models (zero-shot direct, Other-5 systems) into six error types using ground-truth-vs-prediction set arithmetic and relationship-label structure (Table~\ref{tab:cat4_taxonomy}). Of these, 358 return no parseable answer and are reported separately; the shares below are computed over the 642 \emph{substantive} errors. \emph{Default substitution} (51.1\%) and \emph{over-inclusion} (31.9\%) together account for 83.0\% of substantive Cat.~4 errors---both forms of biological-default leakage in which the model fails to apply the cultural override correctly. \emph{Default substitution} is overwhelmingly concentrated in reverse-mapping (Pattern C) questions, accounting for 83.9\% of errors when the model is asked to name the underlying biological relationship behind a cultural kin term. Genuinely wrong-disjoint answers (no overlap with the gold) account for 15.3\% of substantive errors. The 358 empty/format cases sit outside this decomposition entirely; they are concentrated in the reasoning models (DeepSeek/Qwen), and \S\ref{sec:format_robust} shows they are generation-length truncations that do not carry the gap. The 83.0\% biological-default leakage figure is the mechanism signature targeted by the two interventions in \S\ref{sec:interventions}. A parallel manual annotation of 200 CoT responses from GPT-4o-mini and Gemma3-27B yielded a consistent 30.4\% cultural-variation share (cultural-default + over-inclusion), giving a lower-bound corroboration of the automatic estimate.

\begin{table}[h]
\centering
\small
\caption{Cat.~4 error taxonomy on the Other-5 systems (3 OS models, zero-shot direct). Categories are mutually exclusive. Shares are over the 642 \emph{substantive} errors; the 358 predictions with no parseable answer are listed below the rule and excluded from the shares, so that formatting failures are not read as reasoning failures. ``Default sub.'' = bare biological term (uncle, aunt, $\dots$) where the gold requires a paternal/maternal/cross/parallel/classificatory marker; ``over-inclusion'' = pred $\supset$ gold (biological default emitted alongside cultural members); ``partial under'' = pred $\subset$ gold; ``partial overlap'' = mixed; ``wrong / disjoint'' = no overlap with gold. The 51.1\% default substitution rises to 83.9\% on reverse-mapping Pattern C questions.}
\label{tab:cat4_taxonomy}
\begin{tabular}{lrr}
\toprule
\textbf{Error type} & \textbf{N} & \textbf{\% subst.} \\
\midrule
\multicolumn{3}{l}{\textit{Biological-default leakage (83.0\%)}} \\
\quad Default substitution & 328 & 51.1 \\
\quad Over-inclusion       & 205 & 31.9 \\
\midrule
\multicolumn{3}{l}{\textit{Set-shape errors (1.7\%)}} \\
\quad Partial under-coverage & 7 & 1.1 \\
\quad Partial overlap        & 4 & 0.6 \\
\midrule
\quad Wrong / disjoint    & 98  & 15.3 \\
\midrule
\textbf{Substantive Cat.~4 errors} & \textbf{642} & \textbf{100} \\
\midrule
\multicolumn{3}{l}{\textit{Excluded from the shares above}} \\
\quad No parseable answer & 358 & --- \\
\bottomrule
\end{tabular}
\end{table}

\section{Hop Scaling and Stochastic Decoding}
\label{app:hop_models}

Table~\ref{tab:hop_scaling_models} reports the per-model hop curves underlying Figure~\ref{fig:nhops}; the 5- and 6-hop columns are taken from the 8{,}192-token re-run described in \S\ref{sec:hops}. Under the original 512-token budget DeepSeek-R1-32B and Qwen3-32B appeared to collapse to single-digit accuracy at 6 hops. That collapse was an artefact of generation-length truncation rather than a reasoning failure: at the larger budget both models recover to 32 and 52\%. The $\emptyset$ columns report the share of predictions that still return no parseable answer at 8{,}192 tokens; these are scored as errors throughout, so the accuracy columns are a lower bound.

\begin{table}[h]
\centering
\caption{Hop scaling by model (Exact Match \%, zero-shot direct, averaged across systems with data at each depth). The 5- and 6-hop columns use the 8{,}192-token re-run; $\emptyset$5 and $\emptyset$6 give the residual share of predictions with no parseable answer at that budget, which are scored as errors.}
\label{tab:hop_scaling_models}
\small
\setlength{\tabcolsep}{3.4pt}
\begin{tabular}{lcccccccc}
\toprule
\textbf{Model} & \textbf{1} & \textbf{2} & \textbf{3} & \textbf{4} & \textbf{5} & \textbf{6} & \textbf{$\emptyset$5} & \textbf{$\emptyset$6} \\
\midrule
Gemma3-27B & 89 & 92 & 81 & 85 & 41 & 73 & 0 & 0 \\
DeepSeek-R1-32B & 83 & 80 & 71 & 89 & 82 & 32 & 0 & 26 \\
Qwen3-32B & 90 & 84 & 75 & 82 & 65 & 52 & 1 & 17 \\
\midrule
Average & 88 & 85 & 76 & 85 & 63 & 52 & 0 & 14 \\
\bottomrule
\end{tabular}
\end{table}

\begin{figure}[h]
\centering
\includegraphics[width=\columnwidth]{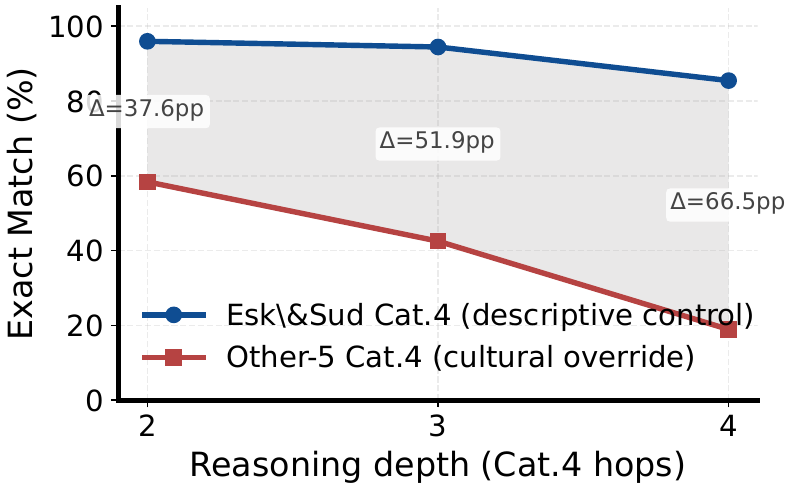}
\caption{Cat.~4 accuracy by reasoning depth on the 50-year ontology (zero-shot direct, mean across the three open-source models; per-cell counts in \S\ref{sec:datasets}). The cultural-override gap between Esk\&Sud (descriptive control) and the Other-5 systems widens monotonically with depth, from 37.6\% at 2-hop to 66.5\% at 4-hop. 5--6 hop Cat.~4 data are reported in Appendix~\ref{app:cat4_56hop}.}
\label{fig:cat4_hop}
\end{figure}

\paragraph{Multi-seed robustness check.}\label{app:multi_seed} The accuracies reported in the main paper use greedy decoding (temperature\,$=0$) for reproducibility, with the $\pm$ values in Tables~\ref{tab:by_system}, \ref{tab:hop_matched}, and \ref{tab:by_system_protocol} reflecting cross-model dispersion across the three open-source models. To confirm that the system-type hierarchy is not a single-trajectory artefact, we evaluate Cat.4 on a stratified subsample of 40 questions per kinship system (280 total) using temperature\,$=0.7$ and $n=5$ samples per question. Each per-question accuracy is the fraction of the 5 samples that exact-match the ground truth.

\begin{table}[h]
\centering
\caption{Multi-seed Cat.4 evaluation (temperature\,$=0.7$, $n=5$ samples, 40-question stratified subsample per system). Mean EM across the three open-source models with $\pm$ cross-model dispersion; per-model bootstrap 95\% CIs in the supplementary repository. The typological ordering of Table~\ref{tab:by_system} is preserved.}
\label{tab:multi_seed_cat4}
\small
\begin{tabular}{llc}
\toprule
\textbf{System} & \textbf{Type} & \textbf{Multi-seed EM (\%)} \\
\midrule
Eskimo    & Descriptive   & 91.0 $\pm$ 5.8 \\
Sudanese  & Descriptive   & 88.7 $\pm$ 10.8 \\
\midrule
Hawaiian  & Generational  & 57.3 $\pm$ 5.1 \\
Iroquois  & Bifurcate     & 43.2 $\pm$ 18.3 \\
Dravidian & Bifurcate     & 61.0 $\pm$ 21.4 \\
Crow      & Mat.\ Skewing & 49.8 $\pm$ 19.1 \\
Omaha     & Pat.\ Skewing & \textbf{33.2 $\pm$ 18.1} \\
\bottomrule
\end{tabular}
\end{table}

Multi-seed accuracies are 3--14\% lower than the greedy numbers in Table~\ref{tab:by_system}, consistent with greedy decoding selecting the modal output (more often correct on this benchmark) and stochastic sampling exposing the underlying decision-noise. Critically, the relative ordering is preserved: descriptive systems (Eskimo, Sudanese) remain $\geq 88\%$, while skewing systems (Crow, Omaha) remain at the bottom, with Omaha still the lowest at 33.2\%. The cross-model dispersion is wider on the bifurcate and skewing systems---further evidence that these are the categories where models genuinely diverge in capability rather than agree on a wrong answer.

\paragraph{5--6 hop Cat.~4 detail.}\label{app:cat4_56hop} Table~\ref{tab:cat4_56hop} reports per-system $\times$ per-model accuracy on the 249 5--6 hop cultural-override questions discussed in \S\ref{sec:hops}. The Other-5 mean EM is 10.6\% (F1 18.9\%), down from 50.4\% at 1--4 hop (Table~\ref{tab:by_system}). The per-system ordering at 5--6 hop is Omaha $>$ Hawaiian $>$ Dravidian $>$ Iroquois $>$ Crow, which differs from the Hawaiian/Dravidian/Iroquois/Crow/Omaha ordering observed at 1--4 hop. Crow's cell ($n=16$) uses six-name set targets; EM is 0\% and F1 is 7.9\%, the latter reflecting partial overlap with the two ground-truth clusters.

\begin{table}[h]
\centering
\caption{5--6 hop Cat.~4 cultural-override accuracy (EM \% / F1 \%, zero-shot direct, max\_tokens\,$=$\,8192). $n$ is the per-system question count. Mean is the per-system unweighted average over the three open-source models.}
\label{tab:cat4_56hop}
\footnotesize
\setlength{\tabcolsep}{3pt}
\resizebox{\columnwidth}{!}{%
\begin{tabular}{lcccccc}
\toprule
\textbf{System} & \textbf{$n$} & \textbf{Gemma3} & \textbf{DeepSeek} & \textbf{Qwen3} & \textbf{EM} & \textbf{F1} \\
\midrule
Hawaiian  & 40 & 35.0/58.4 & \phantom{0}2.5/28.6 & \phantom{0}7.5/13.8 & 15.0 & 33.6 \\
Iroquois  & 80 & \phantom{0}7.5/12.7 & \phantom{0}3.8/\phantom{0}6.3 & \phantom{0}1.2/\phantom{0}2.2 & \phantom{0}4.2 & \phantom{0}7.1 \\
Dravidian & 80 & 17.5/24.3 & \phantom{0}2.5/\phantom{0}5.0 & \phantom{0}5.0/\phantom{0}5.2 & \phantom{0}8.3 & 11.5 \\
Crow      & 16 & \phantom{0}0.0/\phantom{0}7.5 & \phantom{0}0.0/16.2 & \phantom{0}0.0/\phantom{0}0.0 & \phantom{0}0.0 & \phantom{0}7.9 \\
Omaha     & 33 & 36.4/48.3 & 21.2/32.4 & 18.2/22.2 & 25.3 & 34.3 \\
\midrule
\multicolumn{2}{l}{\textit{Other-5 mean}} & 18.5/28.1 & \phantom{0}5.2/13.6 & \phantom{0}5.6/\phantom{0}7.5 & 10.6 & 18.9 \\
\bottomrule
\end{tabular}}
\end{table}

Empty responses (truncation before the answer line) are 0/249 on Gemma3, 5/249 on DeepSeek-R1-32B, and 4/249 on Qwen3-32B, indicating that truncation does not account for the accuracy floor.

\section{Protocol Sensitivity}
\label{app:by_system_protocol}

Table~\ref{tab:by_system_protocol} reports Cat.4 accuracy by kinship system under zero-shot direct and few-shot CoT, supporting the claim in \S\ref{sec:cultural} that the typological ordering persists under enhanced prompting. Few-shot CoT raises Other-5 Cat.4 accuracy modestly on every system, but the descriptive controls (Esk\&Sud) reach near-ceiling, so the system-type gap is preserved.

\begin{table}[h]
\centering
\caption{Performance by kinship system under two protocols (Cat.4 Exact Match \%, mean $\pm$ std across the three open-source models).}
\label{tab:by_system_protocol}
\small
\begin{tabular}{llcc}
\toprule
\textbf{System} & \textbf{Type} & \textbf{Direct} & \textbf{Few-Shot} \\
\midrule
Eskimo & Descriptive & 93.0 $\pm$ \phantom{0}3.3 & 99.6 $\pm$ \phantom{0}0.5 \\
Sudanese & Descriptive & 92.0 $\pm$ \phantom{0}5.2 & 99.5 $\pm$ \phantom{0}0.8 \\
\midrule
Hawaiian & Generational & 61.9 $\pm$ \phantom{0}5.9 & 69.7 $\pm$ \phantom{0}1.9 \\
Dravidian & Bifurcate & 60.4 $\pm$ 13.6 & 63.6 $\pm$ \phantom{0}3.4 \\
Iroquois & Bifurcate & 50.3 $\pm$ 12.1 & 62.4 $\pm$ \phantom{0}7.7 \\
Crow & Mat.\ Skewing & 47.0 $\pm$ 12.8 & 50.0 $\pm$ \phantom{0}9.3 \\
Omaha & Pat.\ Skewing & \textbf{32.6} $\pm$ 11.2 & \textbf{44.4} $\pm$ \phantom{0}4.3 \\
\bottomrule
\end{tabular}
\end{table}

\paragraph{Per-model gap by protocol.}\label{app:gap_by_protocol} Per-model breakdown of the overall (all-category) Esk\&Sud~$\to$~Other-5 gap across protocols, referenced from \S\ref{sec:prompt_robust}.

\begin{table}[h]
\centering
\caption{Overall Esk\&Sud $\to$ Other-5 accuracy gap by protocol (EM \%, mean across all four categories). The gap persists under both protocols for all five models; few-shot CoT raises both sides similarly and shifts the gap by at most a few points in either direction.}
\label{tab:gap_by_protocol}
\setlength{\tabcolsep}{3pt}
\small
\begin{tabular}{llccc}
\toprule
\textbf{Model} & \textbf{Protocol} & \textbf{Esk\&Sud} & \textbf{Other 5} & \textbf{Gap} \\
\midrule
GPT-4o-mini & Direct & 92.3 & 81.2 & 11.1 \\
 & Few-Shot & 98.2 & 82.6 & \textbf{15.5} \\
\midrule
Claude-Haiku-4.5 & Direct & 98.3 & 85.3 & 13.0 \\
 & Few-Shot & 97.6 & 87.9 & \phantom{0}\textbf{9.8} \\
\midrule
Gemma3-27B & Direct & 96.8 & 84.7 & 12.1 \\
 & Few-Shot & 97.6 & 85.0 & \textbf{12.6} \\
\midrule
DeepSeek-R1 & Direct & 92.7 & 73.8 & 18.9 \\
 & Few-Shot & 98.4 & 81.8 & \textbf{16.6} \\
\midrule
Qwen3-32B & Direct & 92.2 & 77.9 & 14.3 \\
 & Few-Shot & 98.2 & 83.1 & \textbf{15.1} \\
\bottomrule
\end{tabular}
\end{table}

\section{Fictional-Rule Control}
\label{app:fictional_control}

To test whether the Cat.4 drop reflects a deficit of cultural \emph{knowledge} or interference from learned cultural \emph{priors}, we construct a relabelled variant of Cat.4 in which the system name and the cultural kin term are replaced with phonetically distant invented labels of identical structural shape. Hawaiian, Iroquois, Dravidian, Crow, and Omaha become Zorblax, Krendari, Glabran, Vespian, and Tarskan; ``classificatory mother/father/sibling/$\dots$'' become ``tier-mater/pater/sibling/$\dots$''; ``cross-cousin'' and ``parallel cousin'' become ``off-axis cousin'' and ``on-axis cousin.'' The biological context, anchor, target persons, ground-truth biological relation, and reasoning depth are preserved exactly; only the cultural-naming surface is changed.

We evaluate the three open-source models on the relabelled Cat.4 questions under zero-shot direct prompting. Table~\ref{tab:fictional_control} reports the per-system, per-model exact-match comparison.

\begin{table}[h]
\centering
\caption{Cat.4 accuracy on real cultural labels vs. fictional isomorphic labels (Exact Match \%, zero-shot direct). Fictional EM exceeds real EM in 11 of 15 cells; the average gain is +6.1\% (95\% bootstrap CI [+2.8, +9.4] over the 15 cells).}
\label{tab:fictional_control}
\setlength{\tabcolsep}{3pt}
\small
\begin{tabular}{llccr}
\toprule
\textbf{Model} & \textbf{System} & \textbf{Real} & \textbf{Fictional} & \textbf{$\Delta$} \\
\midrule
Gemma3-27B       & Hawaiian   & 68.9 & 64.4 & $-$4.5 \\
                 & Iroquois   & 66.7 & 72.0 & +5.3 \\
                 & Dravidian  & 79.5 & 83.3 & +3.8 \\
                 & Crow       & 63.6 & 59.1 & $-$4.5 \\
                 & Omaha      & 47.7 & 53.0 & +5.3 \\
\midrule
DeepSeek-R1-32B  & Hawaiian   & 62.1 & 61.4 & $-$0.8 \\
                 & Iroquois   & 37.9 & 46.2 & \textbf{+8.3} \\
                 & Dravidian  & 51.5 & 65.2 & \textbf{+13.6} \\
                 & Crow       & 32.6 & 30.3 & $-$2.3 \\
                 & Omaha      & 21.2 & 31.1 & \textbf{+9.8} \\
\midrule
Qwen3-32B        & Hawaiian   & 54.5 & 61.4 & +6.8 \\
                 & Iroquois   & 46.2 & 62.9 & \textbf{+16.7} \\
                 & Dravidian  & 50.0 & 63.6 & \textbf{+13.6} \\
                 & Crow       & 44.7 & 53.8 & \textbf{+9.1} \\
                 & Omaha      & 28.8 & 40.1 & \textbf{+11.4} \\
\midrule
\textbf{Average} &            & \textbf{50.4} & \textbf{56.5} & \textbf{+6.1} \\
\bottomrule
\end{tabular}
\end{table}

\paragraph{Interpretation.} If Cat.4 difficulty reflected a pure deficit of cultural \emph{knowledge}, removing the cultural label should leave accuracy unchanged or reduce it (since any helpful prior from training would be lost). Instead, accuracy systematically \emph{rises} when the labels are replaced with invented ones, with the strongest gains on the hardest systems (Iroquois, Dravidian, Omaha) and on the two reasoning-focused models (DeepSeek-R1-32B and Qwen3-32B). This pattern is consistent with a cultural-prior interference account, in which cultural-system labels preferentially activate a learned default (Eskimo-style classification) that competes with the in-context reclassification rule. The orthogonal decomposition below further rules out the simplest lexical/parser explanations. We note that several alternative accounts---altered answer-space expectations, reduced ambiguity in the fictional prompts, or differential parsing behavior around culturally-loaded words---are not directly excluded, and we therefore frame the result as evidence consistent with interference rather than a decisive isolation.

\paragraph{Threats.} The fictional labels (`Zorblax', `Vespian', etc.) may themselves carry weak associations (e.g., reading as fantasy or science fiction); we mitigate by sampling phonotactically distinct strings with no Wikipedia presence, but cannot rule out residual lexical effects. We address the lexical-channel question directly below via an orthogonal decomposition of the two surface swaps.

\paragraph{Orthogonal decomposition (detail).}\label{app:orthogonal_fictional} The summary 2$\times$2 of the system-name vs.\ kin-term swap interaction appears as Table~\ref{tab:fictional_orthogonal} in \S\ref{sec:interventions}. We evaluate the three open-source models under three conditions: \emph{system-only} (only system names swapped, kin terms real), \emph{term-only} (only kin terms swapped, system name real), and \emph{full-fictional} (both swapped). Numbers pool over the $5\times 3 = 15$ (system, model) cells. The interaction is largest for Qwen3-32B (cell-level interactions of +16 to +23\% on most systems); per-cell deltas are in the per-system per-model table above. The model's behaviour does not depend on either surface feature in isolation, but on the joint absence of a recognisable cultural anchor.

\section{In-Context Rule Probe: Full Per-Model Data}
\label{app:rule_context}

Table~\ref{tab:rule_full} reports per-(model, system) baseline and \mbox{+rule} Cat.~4 accuracy on the full Other-5 set (1{,}254 questions per model). Each row is one kinship system; columns give baseline (no rule) and \mbox{+rule} EM/F1 (\%) for each of the three open-source models, plus the per-system mean $\Delta$EM across models. Descriptive controls (Eskimo, Sudanese) are included as a noise-floor sanity check: their prompts are byte-identical between baseline and \mbox{+rule} runs (no rule is injected, since the rule dictionary covers only the Other-5), so any non-zero delta reflects run-to-run non-determinism (0--0\% on Gemma3, $\sim$4\% on DeepSeek, $\sim$7--8\% on Qwen3 under temperature\,$=0$ batched inference). All deltas reported in \S\ref{sec:interventions} exceed this noise floor.

\begin{table}[h]
\centering
\footnotesize
\setlength{\tabcolsep}{3pt}
\caption{In-context-rule probe: per-(model, system) Cat.~4 EM / F1 (\%) under baseline (no rule) vs.\ +rule. Three open-source models, zero-shot direct prompting, \texttt{max\_tokens}\,=\,8192. ``$\Delta$ mean'' is the cross-model average of $\Delta$EM. Empty-response rates (truncation before answer): 0/249 Gemma3, 5/249 DeepSeek-R1-32B, 4/249 Qwen3-32B on the Other-5 cells.}
\label{tab:rule_full}
\resizebox{\columnwidth}{!}{%
\begin{tabular}{lcccccccc}
\toprule
\textbf{System} & \multicolumn{2}{c}{\textbf{Gemma3}} & \multicolumn{2}{c}{\textbf{DeepSeek}} & \multicolumn{2}{c}{\textbf{Qwen3}} & \textbf{$\Delta$ mean} \\
& base & +rule & base & +rule & base & +rule & EM\% \\
\midrule
Eskimo (desc.\ ctrl)   & 96.8/99.4 & 96.8/99.4 & 93.6/95.3 & 97.3/99.5 & 88.8/93.0 & 96.3/97.9 & $+3.7$ \\
Sudanese (desc.\ ctrl) & 98.4/99.7 & 98.4/99.7 & 92.0/95.7 & 97.9/99.4 & 85.6/90.5 & 92.5/95.0 & $+4.3$ \\
\midrule
Hawaiian (Gen.)        & 68.9/76.1 & 57.7/68.2 & 62.1/70.7 & 46.7/58.2 & 54.5/63.7 & 51.1/61.5 & $-10.0$ \\
Iroquois (Bif.)        & 66.7/78.1 & 57.7/71.1 & 37.9/53.4 & 46.7/58.7 & 46.2/62.4 & 52.7/65.8 & $+2.1$ \\
Dravidian (Bif.)       & 79.5/79.5 & 61.5/67.1 & 51.5/61.7 & 53.3/62.1 & 50.0/61.8 & 61.0/68.8 & $-1.7$ \\
Crow (Skewing)         & 63.6/70.0 & 75.6/78.9 & 32.6/45.8 & 46.4/57.2 & 44.7/56.7 & 50.6/61.5 & $+10.6$ \\
Omaha (Skewing)        & 47.7/63.5 & 75.3/79.6 & 21.2/35.1 & 40.4/52.0 & 28.8/41.1 & 51.8/61.0 & $+23.2$ \\
\midrule
\multicolumn{1}{l}{\textit{Other-5 mean}}
                       & 65.3 & 65.6 & 41.1 & 46.7 & 44.8 & 53.4 & $+4.9$ \\
\bottomrule
\end{tabular}}
\end{table}

\paragraph{Per-system replication detail.} The typology\,$\times$\,baseline summary appears as Table~\ref{tab:rule_2x2} in \S\ref{sec:interventions}. Hawaiian is the only non-skewing system where rule supply HURTS on both models with baseline\,$>$\,60\% (Gemma3 $-11.2$, DeepSeek $-15.4$) --- a cross-model replication of the interference signature on a single Generational system. Per-cell deltas above the descriptive noise floor (max $+7.5$\% on Qwen3) are exactly the HELPS/HURTS counts in Table~\ref{tab:rule_2x2}.

\paragraph{Cross-benchmark dissociation.} The baseline column of Table~\ref{tab:rule_full} also exhibits a Gemma3\,$>$\,DeepSeek-R1-Distill-Qwen-32B ordering on Cat.~4 (Other-5 means 65.3\% vs.\ 41.1\%), which reverses the same models' published ordering on standard reasoning benchmarks: on MATH, GPQA, and LiveCodeBench, the reasoning-distilled DeepSeek-R1-Distill leads Gemma3-27B by roughly 5--28\% \citep{deepseek-ai_deepseek-r1_2025,team_gemma_2025}. This dissociation between reasoning-benchmark performance and cultural-rule-application accuracy is consistent with our positioning of Cat.~4 as a capability axis distinct from generic multi-hop reasoning.

\paragraph{Rule statements.} The one-sentence override rule used per system is reproduced in Appendix~\ref{app:prompts}. Rules are derived from standard anthropological descriptions \citep{murdock_social_1949,trautmann_dravidian_1981}; rule-variant ablations (paraphrasing, length, position) are not run here.

\section{Human Baseline}
\label{app:human_pilot}

To establish a human ceiling for KinshipQA and to verify that residual model failures are not artefacts of question ambiguity, we ran a two-annotator human evaluation. We sampled 280 questions stratified by system\,$\times$\,category (10 per cell, 7 systems\,$\times$\,4 categories) and asked two graduate-student annotators with no anthropological background to answer each question from the natural-language context alone. Annotators received the same one-sentence override-rule summary used in the in-context-rule probe (\S\ref{sec:interventions}) for each Other-5 question; Esk\&Sud questions were answered without rule supply. We report per-cell exact-match accuracy and inter-annotator agreement on item-level responses.

\begin{table}[h]
\centering
\small
\setlength{\tabcolsep}{3pt}
\caption{Human baseline per system\,$\times$\,category (\% correct, 10 questions per cell, two annotators A1/A2). Inter-annotator agreement (IAA) on item-level responses is 96.8\% over all 280 questions (per-category: Cat.1 97.1\%, Cat.2 98.6\%, Cat.3 97.1\%, Cat.4 94.3\%).}
\label{tab:human_baseline}
\begin{tabular}{l cc cc cc cc}
\toprule
& \multicolumn{2}{c}{\textbf{Cat.~1}} & \multicolumn{2}{c}{\textbf{Cat.~2}} & \multicolumn{2}{c}{\textbf{Cat.~3}} & \multicolumn{2}{c}{\textbf{Cat.~4}} \\
\cmidrule(lr){2-3} \cmidrule(lr){4-5} \cmidrule(lr){6-7} \cmidrule(lr){8-9}
\textbf{System} & A1 & A2 & A1 & A2 & A1 & A2 & A1 & A2 \\
\midrule
Eskimo    & 100 & 100 & 100 & 100 & 100 & 100 & 100 & 100 \\
Sudanese  & 100 & 100 & 100 & 100 & 100 & 100 & 100 & 100 \\
Hawaiian  & 100 & 100 & 100 & 100 & 100 & 100 & 100 & 100 \\
Iroquois  & 100 & 100 & 100 & 100 & 100 & 100 & 100 & 100 \\
Dravidian & 100 & 100 & \phantom{0}90 & 100 & 100 & 100 & \phantom{0}90 & \phantom{0}90 \\
Crow      & \phantom{0}90 & 100 & 100 & \phantom{0}90 & \phantom{0}90 & \phantom{0}80 & \phantom{0}70 & \phantom{0}80 \\
Omaha     & 100 & \phantom{0}90 & \phantom{0}90 & \phantom{0}90 & \phantom{0}80 & \phantom{0}80 & \phantom{0}80 & \phantom{0}80 \\
\midrule
\textbf{Mean} & \multicolumn{2}{c}{\textbf{98.6}} & \multicolumn{2}{c}{\textbf{97.1}} & \multicolumn{2}{c}{\textbf{95.0}} & \multicolumn{2}{c}{\textbf{90.7}} \\
\bottomrule
\end{tabular}
\end{table}

Humans reach 95.4\% across all 280 questions (annotator means 95.7\% and 95.0\%), with category-level accuracy decreasing along the same hierarchy that LLMs exhibit (Cat.~1: 98.6\%; Cat.~2: 97.1\%; Cat.~3: 95.0\%; Cat.~4: 90.7\%). On the cultural-override category (Cat.~4), human accuracy on the Other-5 systems averages 89.0\% (annotator means 88\% and 90\%), compared to the open-source LLM mean of 50.4\% (Table~\ref{tab:by_system}) --- a 38.6\% human--model gap. The per-system pattern within Cat.~4 also matches the typological ordering: humans saturate the descriptive controls and the generational/bifurcate systems where the override is structurally simple (100\% on Eskimo, Sudanese, Hawaiian, Iroquois) and drop on the skewing systems (Crow 75\%, Omaha 80\%) and on the marriage-rule-constrained Dravidian (90\%). The two skewing systems are also the lowest-scoring cells under inter-annotator agreement, consistent with their structural difficulty rather than question ambiguity. Multilingual native-speaker validation is left to future work.

\section{Frontier-scale Ceiling Check}
\label{app:opus_check}

To test whether the cultural-override pattern persists at frontier scale, we evaluate \textbf{Claude Opus 4.7} (\texttt{claude-opus-4-7}, Anthropic's flagship reasoning-augmented model) on a stratified Cat.~4 subsample: 100 questions per system (700 total) drawn with seed 42 across the same dataset used for the open-source evaluation, under the same zero-shot direct prompt template (Appendix~\ref{app:prompts}) and parser. Greedy decoding (temperature\,$=0$) is used where supported by the API.

\begin{table}[h]
\centering
\caption{Frontier-scale Cat.~4 spot-check: Claude Opus 4.7 (zero-shot direct and few-shot CoT) vs.\ the three open-source models. Open-source numbers are per-system Cat.~4 EM under zero-shot direct from Table~\ref{tab:by_system_protocol}; the Opus 4.7 columns use 100 stratified Cat.~4 questions per system. The typological ordering of Table~\ref{tab:by_system} is preserved at frontier scale; few-shot CoT narrows the cultural-override gap but does not close it.}
\label{tab:opus_check}
\footnotesize
\setlength{\tabcolsep}{2pt}
\begin{tabular}{llcccc}
\toprule
\textbf{System} & \textbf{Type} & \textbf{Open-src} & \textbf{Dir.} & \textbf{FS-CoT} & \textbf{$\Delta_{\text{FS}}$} \\
\midrule
Eskimo    & Descriptive   & 93.0 & \textbf{100.0} & \textbf{100.0} & \phantom{$-$0}$+0$ \\
Sudanese  & Descriptive   & 92.0 & \textbf{100.0} & \textbf{100.0} & \phantom{$-$0}$+0$ \\
\midrule
Hawaiian  & Generational  & 61.9 & 74.0 & 84.0 & \phantom{0}$+10$ \\
Iroquois  & Bifurcate     & 50.3 & 70.0 & 75.0 & \phantom{0}$+5$ \\
Dravidian & Bifurcate     & 60.4 & 87.0 & 84.0 & \phantom{0}$-3$ \\
Crow      & Mat.\ Skew.\  & 47.0 & 60.0 & 76.0 & $+16$ \\
Omaha     & Pat.\ Skew.\  & 32.6 & 54.0 & \textbf{73.0} & $+19$ \\
\midrule
\multicolumn{2}{l}{\textit{Esk\&Sud avg.}}  & 92.5 & \textbf{100.0} & \textbf{100.0} & \phantom{0}$+0$ \\
\multicolumn{2}{l}{\textit{Other-5 avg.}}   & 50.4 & 69.0 & 78.4 & \phantom{0}$+9.4$ \\
\multicolumn{2}{l}{\textbf{Cultural-override gap}} & \textbf{42.1} & 31.0 & \textbf{21.6} & $-9.4$ \\
\bottomrule
\end{tabular}
\end{table}

Three observations sharpen the paper's central claim. \emph{First,} frontier-scale reasoning lifts Cat.~4 accuracy by $+18.6$\% on average across the Other-5 systems and saturates the descriptive controls (Eskimo and Sudanese both reach 100\%); the cultural-override gap narrows from 42.1\% to 31.0\% under zero-shot direct. \emph{Second,} few-shot CoT at frontier scale further narrows the gap to 21.6\%---about half of the open-source baseline closes when scale and demonstrations are combined---with the largest per-system gains on the hardest systems (Omaha $+19$, Crow $+16$). \emph{Third,} the gap nevertheless persists: even Opus 4.7 with few-shot CoT misses 16--27\% of Cat.~4 questions on the bifurcate, generational, and skewing systems, and the per-system relative ordering is fully preserved. The cultural-override phenomenon is therefore not a mid-tier-only artefact: it persists at frontier scale, is partially mitigated by combining scale with few-shot demonstrations, and shows a stable cross-system structure.

\paragraph{Two-factors-compound pattern at frontier scale.} Binning the Opus 4.7 spot-check by reasoning depth (Table~\ref{tab:opus_by_hop}) shows the cultural-override gap also \emph{widens with hop count} at frontier scale---from 22.4\% at 2-hop to 47.6\% at 3-hop, a +25.2\% widening. The open-source models widen by a comparable amount in absolute terms (+17.4\% from 37.1$\to$54.5\%), indicating that the depth\,$\times$\,override interaction documented in \S\ref{sec:hop_matched} for the open-source models is not eliminated by frontier-scale reasoning capacity.

\begin{table}[h]
\centering
\caption{Cat.~4 cultural-override gap by reasoning depth, Opus 4.7 vs. the three open-source models. The gap widens with hop in both, with similar absolute widening ($\sim\!17$--$25$\%).}
\label{tab:opus_by_hop}
\small
\begin{tabular}{llccc}
\toprule
\textbf{Cohort} & \textbf{Hop} & \textbf{Esk\&Sud} & \textbf{Other-5} & \textbf{Gap} \\
\midrule
Opus 4.7      & 2 & 100.0 & 77.6 & 22.4 \\
              & 3 & 100.0 & 52.4 & \textbf{47.6} \\
\midrule
Open-source   & 2 & \phantom{0}95.7 & 58.6 & 37.1 \\
(3-model avg) & 3 & \phantom{0}96.7 & 42.2 & \textbf{54.5} \\
\bottomrule
\end{tabular}
\end{table}

\section{Paraphrase Robustness Probe}
\label{app:paraphrase_robust}

To test whether the cultural-override gap reflects sensitivity to the specific question template used in the headline benchmark (a possible ``phrasing artefact'' explanation), we re-run Gemma3-27B (the highest-accuracy open-source model on Cat.~4) on a 50-question Cat.~4 subsample (10 per Other-5 system, seed 20260523, stratified across 2-hop and 3-hop) with the question text replaced by a rule-paraphrased variant.

\paragraph{Paraphrase construction.} Paraphrases preserve entity names, the system identifier, and the kin term verbatim, and vary only the surrounding framing (Table~\ref{tab:paraphrase_examples}). Patterns cover the four most common Cat.~4 question shapes in the dataset: forward-singular (``In the X kinship system, who is A's KT?''), forward-plural (``In the X kinship system, who are A's KTs?''), forward-list (``List all of A's KTs according to the X system''), and reverse-mapping (``In the X system, A is called B's `KT'. What is their actual biological relationship?''). The context paragraph (NL serialisation of the proof subgraph) is held identical to the original.

\begin{table}[h]
\centering
\small
\caption{Example paraphrases (one per pattern). All four templates preserve entity names, system label, and kin term verbatim.}
\label{tab:paraphrase_examples}
\setlength{\tabcolsep}{3pt}
\begin{tabular}{p{0.46\columnwidth} p{0.46\columnwidth}}
\toprule
\textbf{Original} & \textbf{Paraphrase} \\
\midrule
In the Hawaiian kinship system, who is Lisa W.'s classificatory mother? & Under the Hawaiian kinship convention, identify Lisa W.'s classificatory mother. \\
In the Iroquois kinship system, who are M.W.'s cross-cousins? & Under the Iroquois kinship convention, enumerate M.W.'s cross-cousins. \\
List all of P.J.'s classificatory siblings according to the Hawaiian system. & Within the Hawaiian system, give the complete list of P.J.'s classificatory siblings. \\
In the Hawaiian system, M.W.\ is called D.J.'s `classificatory mother'. What is their actual biological relationship? & Per Hawaiian kinship conventions, M.W.\ holds the role of `classificatory mother' relative to D.J. State their actual biological relationship. \\
\bottomrule
\end{tabular}
\end{table}

\paragraph{Results.} Table~\ref{tab:paraphrase_robust} reports Gemma3-27B EM on the 50 originals (anchor) and on the paraphrased variants. Aggregate EM rises from 78.0\% to 84.0\% ($+6.0$\%); F1 rises from 87.1\% to 88.3\% ($+1.2$\%). Five questions flip from incorrect to correct under paraphrase and two flip the other way, leaving net $+3$ correct out of 50. Per-system, four of five systems are stable or improve (Hawaiian, Iroquois +10\%; Omaha +20\%; Dravidian unchanged at 100\%); Crow drops 10\% (one question, $n=10$). The direction of change is the opposite of what a brittle template-artefact account would predict; the absence of degradation is consistent with the Cat.~4 cultural-override gap reflecting a cultural-knowledge axis rather than dependence on the specific phrasing used in our headline benchmark.

\begin{table}[h]
\centering
\small
\caption{Paraphrase robustness on Gemma3-27B (zero-shot direct, $n=10$ per system, total $n=50$). EM is computed against the same gold answers used in the headline runs. Net flips $=$ (orig-wrong, para-right) minus (orig-right, para-wrong).}
\label{tab:paraphrase_robust}
\begin{tabular}{lccc}
\toprule
\textbf{System} & \textbf{Orig EM} & \textbf{Para EM} & \textbf{$\Delta$} \\
\midrule
Hawaiian  & 90.0 & 100.0 & $+10.0$ \\
Iroquois  & 90.0 & 100.0 & $+10.0$ \\
Dravidian & 100.0 & 100.0 & \phantom{0}$+0.0$ \\
Crow      & 70.0 & \phantom{0}60.0 & $-10.0$ \\
Omaha     & 40.0 & \phantom{0}60.0 & $+20.0$ \\
\midrule
\textbf{Other-5 avg.} & \textbf{78.0} & \textbf{84.0} & $+6.0$ \\
\textbf{Net flips}    & \multicolumn{3}{c}{$+5$ wrong$\to$right, $-2$ right$\to$wrong} \\
\bottomrule
\end{tabular}
\end{table}

\paragraph{Caveats.} The probe covers a single open-source model ($n=1$) on a single stratified subsample ($n=50$); the $+6.0$\% aggregate is within sampling noise for $n=50$, and the per-system entries should not be over-interpreted at $n=10$ per cell. The probe is a controlled-direction stress test against the template-artefact hypothesis, not a broad robustness study. Paraphrase generation is rule-based via four templates (one per question pattern), which avoids LLM-paraphraser contamination but limits the surface diversity probed. Full paraphrase set and per-question predictions are released alongside the dataset.

\end{document}

%% file: latex/introduction-Dimitar.tex
Large language models (LLMs) are increasingly evaluated not only on their ability to recall isolated facts, but also on their capacity to combine multiple pieces of information into a coherent answer. A particularly challenging instance of this broader goal is multi-hop reasoning: the ability to infer a correct conclusion by chaining together several intermediate relations, including such that are not explicitly stated. In this paper, we propose a novel way to test this ability by making use of data about kinship relations.

Our approach is based on producing computer-generated realistic datasets representing a collection of interconnected family trees constructed to comply with a set of constraints regulating marriage in a given society. These constraints, often perceived as cultural or legal `taboos', determine which kin relations are possible, forbidden, or socially marked. By controlling these constraints, we can generate large, internally consistent genealogical datasets tailored to specific cultural settings.

Family trees are a particularly suitable domain for evaluating multi-hop reasoning. First, kinship is a natural and ubiquitous topic of human communication, and kin vocabulary belongs to the most stable layers of language \cite{Swadesh52}. Second, genealogical structures offer a controlled way to vary task complexity: simple relations can be defined over one or two edges in a tree, while more complex relations require chaining over many intermediate nodes. This makes kinship an ideal testbed for probing how well an LLM can integrate multiple facts into a single inference.

Many cultures and their languages lexicalise kin relations that span several nodes in a family tree (e.g.\ Bulgarian {\fontfamily{cmr}\itshape\foreignlanguage{bulgarian}{девер}}, denoting one's husband's brother), letting speakers refer succinctly to specific relatives without spelling out the intermediary chain. Conversely, in some societies the closest translation of a kin term may refer not only to a biological mother but also to maternal aunts. Despite this cross-linguistic variation, any biological kin relation can in principle be described from elementary notions such as biological father and mother; what varies is the ease, compactness, and conventionality of such descriptions.

These observations suggest that LLM performance on kinship inference tasks may vary depending on the language and culture involved, and on the model’s implicit knowledge of the relevant kinship vocabulary and concepts. Nevertheless, the task itself remains well defined: inferring the relationship between two individuals from a set of stated facts. As such, it provides a robust and interpretable way to assess multi-hop reasoning ability.

This paper makes the following contributions:
\begin{itemize}\itemsep0pt\parsep0pt
\item \textbf{A procedurally-extensible kinship benchmark generator}: a tunable simulator horizon (50-, 150-, and 200-year) scales population size and chain depth on demand, eliminating exact-instance overlap with pretraining data.
\item \textbf{KinshipQA}: a 3{,}931-question benchmark across seven anthropologically-documented kinship systems, four reasoning categories, and reasoning depths from 1 to 6 hops.
\item \textbf{Cultural disambiguation imposes a robust additional burden beyond multi-hop chaining.} We document a 40.9\% Cat.~1--3~$\to$~Cat.~4 EM drop on the five non-descriptive systems; the drop holds for every one of them and is largest for the two skewing systems (Crow, Omaha). It persists under chain-of-thought, few-shot prompting, and depth-matched comparison, and compounds with chain depth, falling to 10.6\% at 5--6 hops while biological composition over the same chains holds at 58.6\%. A two-annotator human baseline ($n=280$, IAA 96.8\%) reaches 89.0\% on this category under matched in-context rule supply; LLMs given the same rule reach only 50.7\%, so the questions are reliably solvable once the rule is supplied.
\item \textbf{Two interventions on the Cat.~4 gap.} A fictional-rule control replaces system labels and kin terms with invented strings and raises Cat.~4 accuracy by 6.1\%; an in-context-rule probe prepends the override rule and produces opposite-signed effects across typology, reducing accuracy on non-skewing systems with high baseline and raising it on skewing systems. The two interventions together help distinguish surface-form effects from a possible missing skewing prior; per-(model, system) results in App.~\ref{app:rule_context} and the orthogonal 2$\times$2 label/kin-term decomposition in App.~G.
\end{itemize}